\documentclass{article}

\usepackage{microtype}
\usepackage{graphicx}
\usepackage{subfigure}
\usepackage{booktabs} % for professional tables
\usepackage{amsmath}
\usepackage{amssymb}
\usepackage{caption}
\usepackage{multirow, multicol}
\usepackage{enumitem}

\usepackage{hyperref}

\usepackage[accepted]{icml2021}

\icmltitlerunning{Test-time Adaptation for Real Image Denoising via Meta-transfer Learning}

\begin{document}

\twocolumn[
\icmltitle{Test-time Adaptation for Real Image Denoising via Meta-transfer Learning}



\icmlsetsymbol{equal}{*}

\begin{icmlauthorlist}
\icmlauthor{Agus Gunawan}{equal,ka}
\icmlauthor{Muhammad Adi Nugroho}{equal,ka}
\icmlauthor{Se Jin Park}{equal,ka}
\end{icmlauthorlist}

\icmlaffiliation{ka}{Department of Electrical Engineering, Korea Advanced Institute of Science and Technology, Korea}

\icmlcorrespondingauthor{Agus Gunawan}{agusgun@kaist.ac.kr}
% \icmlcorrespondingauthor{Eee Pppp}{ep@eden.co.uk}

\icmlkeywords{Machine Learning, ICML}

\vskip 0.3in
]



\printAffiliationsAndNotice{\icmlEqualContribution} % otherwise use the standard text.

\begin{abstract}
In recent years, a ton of research has been conducted on real image denoising tasks.
However, the efforts are more focused on improving real image denoising through creating a better network architecture.
We explore a different direction where we propose to improve real image denoising performance through a better learning strategy that can enable test-time adaptation on the multi-task network.
The learning strategy is two stages where the first stage pre-train the network using meta-auxiliary learning to get better meta-initialization.
Meanwhile, we use meta-learning for fine-tuning (meta-transfer learning) the network as the second stage of our training to enable test-time adaptation on real noisy images.
To exploit a better learning strategy, we also propose a network architecture with self-supervised masked reconstruction loss.
Experiments on a real noisy dataset show the contribution of the proposed method and show that the proposed method can outperform other SOTA methods.
\end{abstract}

\section{Introduction}
Image noise can cause performance degradation on various tasks \cite{Koziarski2017image}. 
To solve this, image denoising techniques are developed for image recovery. 
One of the early deep learning approaches for denoising is \cite{zhang2017beyond} that propose DnCNN, a residual learning strategy to solve AWGN denoising tasks. 
However, simple statistic AWGN-based noise cannot model real-world noise. 
This encourages developments toward denoising real noisy images, where the key objective is to create a network that can adapt to various types of noises or has better generalization so it can work across various types of noises \cite{Lin2019real,kim2019grdn,tian2021dualcnn}.

In the case of adapting to various types of degradations in various low-level vision tasks, several methods use an optimization-based meta-learning paradigm to enable test-time adaptation.
\cite{chi2021test} proposes to use meta-auxiliary learning for fine-tuning to enable test-time adaptation in deblurring. 
While, \cite{lee_2020} and \cite{soh2020meta} propose to use meta-learning for fine-tuning i.e. meta-transfer learning \cite{sun2019meta} to enable test-time adaptation in denoising and super-resolution tasks respectively.
Different from previous works in various aspects, our methods apply meta-auxiliary learning in pre-training and use meta-transfer learning to achieve better generalization and enable test-time adaptation.
We design two networks which are mask generation network and multi-task network. 
Our goal is to utilize the two stages of learning for the multi-task network such that when the head of the multi-task network is updated using auxiliary loss, the denoising performance can be improved in any dataset.
We use self-supervised masked reconstruction loss as the auxiliary loss and the mask is generated by the mask generation network. 

The motivation of using masked reconstruction loss is to encourage the auxiliary head to produce only the noisy part of the image that can benefit the primary task.
% By encouraging the auxiliary head to produce the noisy part of the image, the primary head also gains some benefit since it knows where is the noisy part of the image and which region that the primary head needs to denoise.
Our motivation comes from various literature such as: 1) \cite{zhang2017beyond} that shows better performance when the network is trained with noise ground truth instead of a clean image ground truth, and 2) \cite{yang2017ct} that shows the loss of high-frequency components due to over smoothing when trained using only reconstruction loss between predicted clean image and clean image ground truth. 
When the network knows the region of the noisy part of the image, it can focus more on that instead of denoising other unrelated parts which can make the region become over smooth.
This problem also becomes important for real noises cases since \cite{zhou2020awgn} shows that real noises are mostly spatially/channel-correlated and
spatially/channel-variant.

Furthermore, we meta-learn the mask generation network in two stages.
The first stage uses meta-auxiliary learning to encourage the mask generation network to produce a mask that can improve the generalization of the multi-task network's primary task against various types of synthetic noise when trained using the auxiliary objective.
Meanwhile, the second stage uses meta-transfer learning to make the mask generation network produce a mask that can benefit primary tasks of the multi-task network against real noises.
The produced mask will enable test-time adaptation of multi-task network when multi-task network is trained using masked reconstruction loss, which will improve the performance of denoising task in the corresponding dataset without any ground truth.

The contributions of our paper are as follows:

\begin{itemize}[parsep=0pt,partopsep=-5pt]
    \item We design a network architecture that can gain more improvements on the primary task when trained using the auxiliary objective.
    \item We propose masked reconstruction loss as an auxiliary objective to improve the denoising task. Note that our masked reconstruction loss also may be used in other low-level vision tasks such as super-resolution or deblurring.
    \item We propose to use the meta-auxiliary learning method to pre-train the network and use meta-transfer learning to make the network can adapt across various types of noise and enable test-time adaptation. In addition, we only update the heads of the multi-task network to enable faster adaptation.
\end{itemize}

\section{Related Work}
\subsection{CNN-based Image Denoising}
Image denoising is recovering a clean image $x$ from a noisy image $y$  that follows an image degradation model $y = x + n$. 
The common assumption is that the noise $n$ is an additive white Gaussian noise (AWGN).
With the recent advances in deep learning, numerous deep learning-based methods have been proposed \cite{zhang2017beyond, zhang2017learning, liu2018non, zhang2018ffdnet, zhang2019residual, zhang2020residual}. 
DnCNN \cite{zhang2017beyond} exploits a deep neural network to speed up training and boost the performance with residual learning. 
FFDNet \cite{zhang2018ffdnet} takes cropped images and a noise level map to handle locally varying and different ranges of noise levels. 
RNAN \cite{zhang2019residual} is a residual non-local attention network that can consider long-range dependencies among pixels.
RIDNet \cite{anwar2019real} uses residual-in-residual structure to help low-frequency information flows and uses feature attention to exploit channel dependencies.
RDN \cite{zhang2020residual} is a deep residual dense network that can extract hierarchical local and global features. 
MIRNet \cite{zamir2020learning} design a novel network architecture to maintain spatially-precise high-resolution representations and strong contextual information in the entire network by using a multi-scale residual block with residual connection and attention mechanism.
However, they rely on a large number of training datasets with paired noisy and ground truth clean images and highly depend on the distribution of the training data. 
The same set of training weights are used for test images, thereby failing under the distribution shift of the data. 
To overcome this limitation, zero-shot denoising has been proposed to learn image-specific internal structure. 

\subsection{Zero-shot Denoising}
Zero-shot denoising aims to denoise images on the zero-shot setting to be easily adapted to the test image condition. 
To be less affected by the noise distribution of the training data, several works have proposed to train without true clean images with the assumption of zero-mean noise. 
Noise2Noise \cite{lehtinen2018noise2noise} trains with pairs of noisy patches and is based on the reasoning that the expectation of the randomly corrupted signal is close to the clean image. 
Noise2Void \cite{krull2019noise2void} only considers the center pixel of the input patch and is trained to predict the center pixels.
However, they do not exploit the large-scale external dataset and therefore show inferior performance compared to supervised methods where the distribution of the test input is identical to the training data distribution. 
Different from previous methods, our method exploits the large-scale external dataset by training the method using meta-auxiliary learning.
Then, we enable test-time adaptation by training the network using meta-learning with the help of self-supervised auxiliary loss so it can learn image-specific internal structure.

\subsection{Meta-learning and Meta-auxiliary Learning}
Meta-learning aims to learn new concepts quickly with a few examples by learning to learn. 
In this respect, meta-learning is considered together with few-shot and zero-shot learning. 
Meta-learning is categorized into three groups; metric-based, memory network-based, and optimization-based.
Among them, MAML \cite{MAML_finn} which is one of the optimization-based methods has shown a great impact on the research community. 
Meta-learning has two phases; meta-training and meta-test. 
In meta-training, a task is sampled from a task distribution and training samples are used to optimize the base-learner with a task-specific loss, and test training samples are used to optimize the meta-learner. 
In the meta-test, the model adapts to a new task with the meta-learner.
\cite{MAML_finn} adopts a simple gradient descent algorithm to find an initial transferable point where a few updates can fast adapt to a new task. 
ZSSR \cite{soh2020meta} additionally leverages meta-transfer learning for zero-shot super resolution (MZSR). 
Auxiliary learning has been integrated with meta-learning, so-called Meta AuXiliary Learning (MAXL) \cite{liu2019self}. 
MAXL consists of a label-generation network to predict the auxiliary labels, and a multi-task network to train the primary task and the auxiliary task. 
The interaction between the two networks is a form of meta-learning with a double gradient. 
As the auxiliary task is self-supervised, it has down promising direction in zero-shot meta-learning. 
MaXLDeblur \cite{chi2021test} incorporates meta-auxiliary learning for transfer learning in dynamic scene deblurring task. 
It uses a self-reconstruction auxiliary task that shares layers with the primary deblurring task, which gains performance via the auxiliary task.
The model is adapted to each input image to better capture the internal information, thereby allowing fast test-time adaptation.
One related work that applies meta-learning in denoising tasks is the work from \cite{lee_2020}.
This work proposes self-supervised loss coupled with meta-learning to enable adaptation in test-time.
However, this method and the self-supervised loss only works for synthetic noise and cannot be applied to real-world noise case.
In our paper, we apply meta-auxiliary learning for pre-training and meta-learning for fine-tuning (meta-transfer learning) in image denoising problems that will enable adaptation in test-time and achieve better generalization. 
Our meta-learning problem is similar to \cite{chi2021test} which can be seen as zero-shot meta-learning that tries to make the network can fast adapt to specific noise in one image by using a few updates of auxiliary loss.

\section{Proposed Methods}

\begin{figure*}[ht]
\vskip 0.1in
\begin{center}
    \centerline{\includegraphics[width=0.65\textwidth]{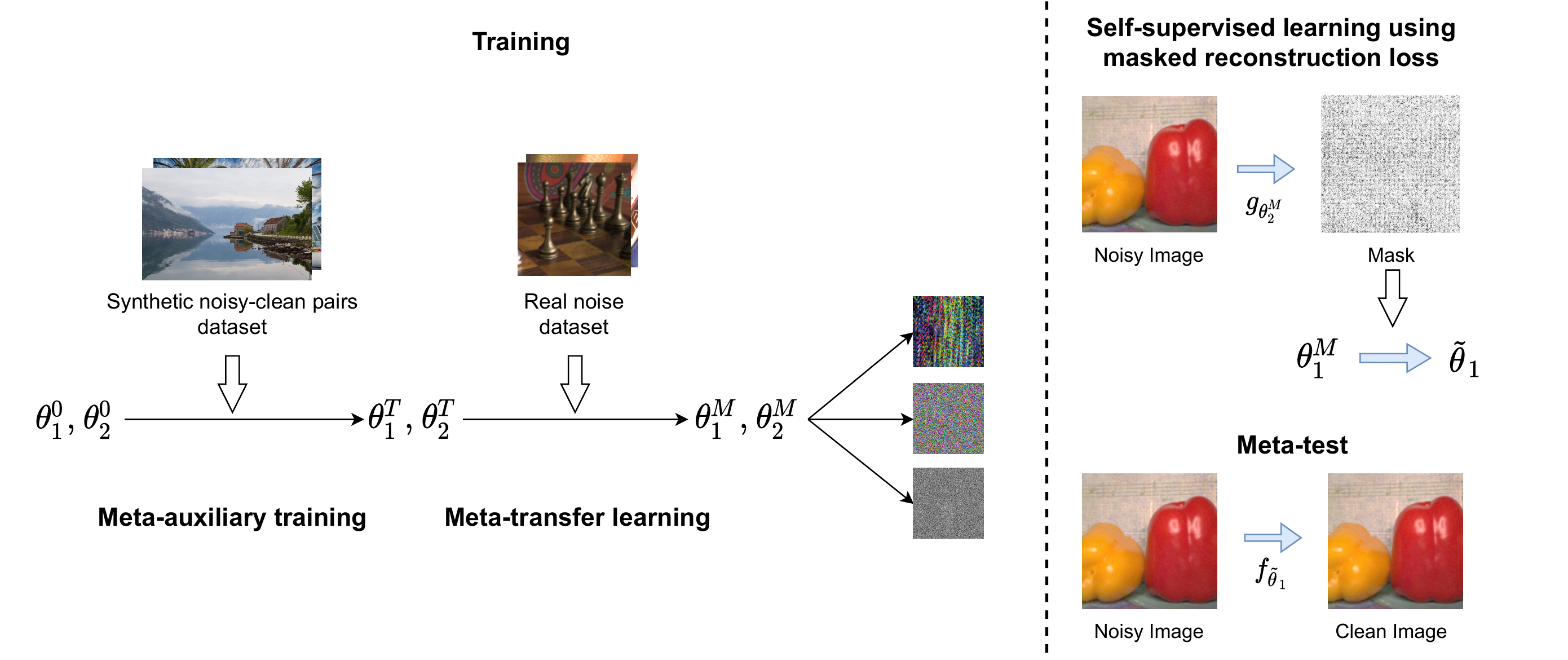}}
    \caption{The overview of our learning method. Our network (multi-task and mask generation network) is trained from random initialization $\theta_{1}^{0}, \theta_{2}^{0}$ to $\theta_{1}^{T}, \theta_{2}^{T}$ using meta-auxiliary learning. Then, we use meta-transfer learning to learn representation $\theta_{1}^M, \theta_{2}^M$, where $\theta_{1}^{M}$ will have a good representation to denoise various noise models and the performance will be improved if trained using masked reconstruction loss. Then, for each test image, we adapt the denoising network using self-supervised masked-reconstruction loss.}
    \label{fig:method-overview}
    \end{center}
    \vskip -0.2in
\end{figure*}

Given a noisy image $I_{n}$, our network (multi-task branch) output predicted clean image $\hat{I}_{c}$ and predicted noisy image $\hat{I}_{n}$.
In addition, the mask generation branch $g_{\theta_2}$ of our network also produces a mask $M$ to condition the reconstruction loss $L_{Rec}$ that is used as an auxiliary loss $L_{Aux}$ to train our multi-task network $f_{\theta_1}$. 
The overview of our method can be seen in Figure \ref{fig:method-overview}.
First, we train the multi-task network and mask generation network using meta-auxiliary learning to provide better meta-initialization.
This is because meta-auxiliary learning can improve the generalization of the network for robustness against various synthetic noises.
% The main focus in this stage is to make the mask produced by the mask generation network can improve the primary task performance (denoising).
Then, we use this pre-train network as meta-initialization for the meta-transfer learning.
In this stage of learning, the objective is to make the multi-task network can improve the primary task performance when the parameter of the network is updated by auxiliary loss (i.e. masked reconstruction loss) in real noises cases.
In addition, using these two stages of learning, we want to make the mask generation network produce a better mask that will help the multi-task network can adapt to various types of noises (synthetic and real) when trained using masked reconstruction loss.
Then, for the test dataset of unseen data, we adapt the parameter of the multi-task network on each image example (i.e. zero-shot meta-learning) by using masked reconstruction loss which can be trained in a self-supervised manner without any ground truth.

\subsection{Network Architecture}

\begin{figure*}[ht]
\vskip 0.1in
\begin{center}
    \centerline{\includegraphics[width=0.6\textwidth]{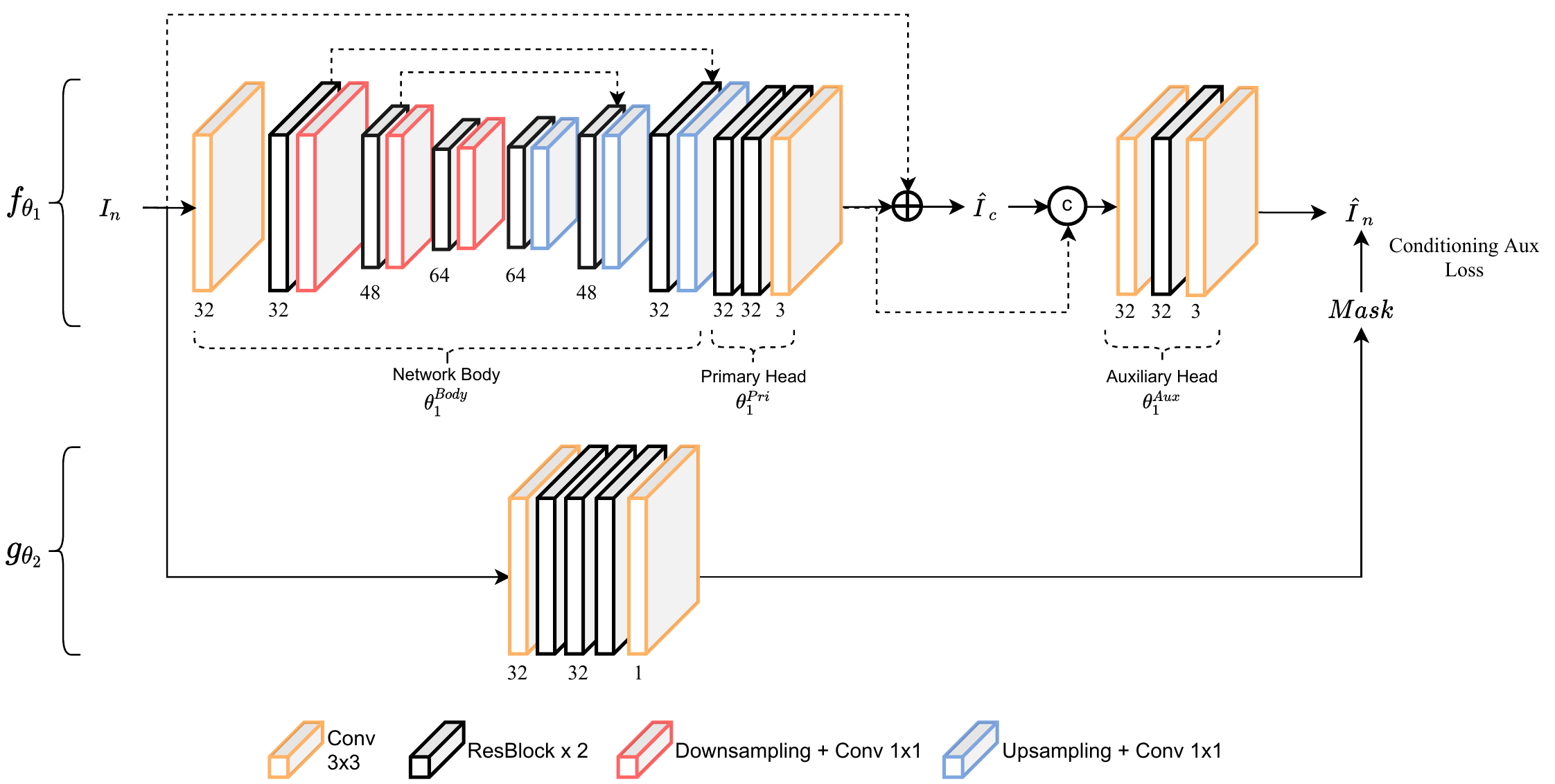}}
    \caption{The architecture of our network.}
    \label{fig:network-architecture}
    \end{center}
    \vskip -0.2in
\end{figure*}

Inspired by \cite{liu2019self}, we design a network that consists of a multi-task network and a mask generation network. 
The architecture of the network can be seen in Figure \ref{fig:network-architecture}. 
The multi-task network goal is to solve two tasks which are denoising (primary task) and noisy image prediction (auxiliary task). 
In the multi-task network, we use a single convolution layer and encoder-decoder with skip connection as the network body. 
The network body will produce deep features which will be used by the primary head to refine the feature resulting in the residual image. 
This residual image when added with the noisy image $I_{n}$ will produce the predicted clean image $\hat{I}_{c}$.

After that, we concatenate the predicted clean image and the residual.
The auxiliary head will use this concatenation to produce the predicted noisy image $\hat{I}_{n}$.
We design the network so the auxiliary head uses the output of the primary head and predicted clean image. 
This is intended since we will only train the primary head and auxiliary head in the inner loop of the meta-transfer learning and meta-testing step for test-time adaptation.
The rationale behind not updating the network body is because \cite{raghu2020rapid} shows MAML-based optimization produces only a little change on the network body parameters. 
As an effect, we can do fast test-time adaptation with less memory and computation in the meta testing when encountered with unseen data.
In addition, we also observe that placing the auxiliary head after the primary head gains more benefits compared to a single feature extractor with multi-head architecture when trained with auxiliary loss (Section \ref{subsec:ablation-study}).

The loss to train the multi-task network consists of auxiliary loss $L_{Aux}$ and primary loss $L_{Pri}$.
The primary loss $L_{Pri}$ is the reconstruction loss between predicted clean image $\hat{I}_{c}$ and clean ground truth image $I_{GT}$ which can be formulated as:
\begin{equation}
    \label{eq:primary-loss}
    L_{Pri}(\hat{I}_{c}, I_{GT}) = \lvert \lvert \hat{I}_{c} - I_{GT} \rvert \rvert_{1}
\end{equation}

Meanwhile, we use masked reconstruction loss $L_{MaskRec}$ as the auxiliary loss $L_{Aux}$ which can be formulated as:
\begin{equation}
    \label{eq:aux-loss}
    \begin{aligned}
    L_{Aux}(\hat{I}_{n}, I_{n}, Mask) &= L_{MaskRec}(\hat{I}_{n}, I_{n}, Mask)\\
    &= L_{Rec}(\hat{I}_{n}, I_{n}) \odot Mask\\ 
    &=\lvert \lvert \hat{I}_{n} - I_{n} \rvert \rvert_{1} \odot Mask
    \end{aligned}
\end{equation}
In this auxiliary loss $L_{Aux}$, we conditioned the reconstruction loss $L_{Rec}$ between the predicted noisy image $\hat{I}_{n}$ and the noisy image $I_{n}$ on the mask produced by the mask generation network. 
By doing this, we only compute auxiliary loss on some pixels that will improve the primary task performance.
In addition, this auxiliary loss is also self-supervised since it does not require any ground truth that makes this loss appropriate to be applied in the test time.
% The mask generation network will be trained based on the primary task performance both on pre-training using meta-auxiliary learning and meta-transfer learning stage.

\subsection{Pre-training using Meta-auxiliary Learning (MAXL)}

Similar to \cite{chi2021test, soh2020meta}, we train our network using an external dataset.
However, our method trains the network using a meta-auxiliary learning scheme similar to \cite{liu2019self}.
The goal of using this scheme is to improve the generalization power of our network that can serve as a better meta initialization before doing meta-transfer learning. 
In addition, we find that using masked reconstruction loss only can make the mask collapse (one or zero at every pixel).
To solve this issue, we use two-directional image gradient loss \cite{zhang2018densely} to regularize the mask.
By using this loss for regularizing the mask, we can force the mask generation network to produce a mask that has a similar edge with the noisy image.
As an effect, it may help the multi-task network to denoise the image better and prevent the loss of fine-textural details which is a common issue in denoising networks \cite{anwar2019real, zamir2020learning}.
The two-directional gradient loss can be formulated as:
\begin{equation}
    \label{eq:two-drirectional-gradient-loss}
    \begin{aligned}
    L_G(M, I_n) = &\sum_{w,h}\lvert\lvert(H_x(M))_{w,h} - (H_x(I_n))_{w,h}\rvert\rvert \\
    & + \lvert\lvert(H_y(M))_{w,h} - (H_y(I_n))_{w,h}\rvert\rvert
    \end{aligned}
\end{equation}

where $M$,$w$,$h$ are respectively mask, width, and height. 
$H_x$ and $H_y$ are image gradient operators along rows (horizontal) and columns (vertical).
However, we only use this loss in the pre-training stage.
This is because in the meta-transfer learning stage the collapse issue does not appear.
Moreover, the mask always changes through learning which denotes that the multi-task network needs to reconstruct different regions of the noisy image to help in improving the denoising performance (Figure \ref{fig:mask-maxl-mtl} and Figure \ref{fig:mask-scratch-mtl} in supplementary).
The algorithm to train the network follows the algorithm from \cite{liu2019self} in Algorithm \ref{alg:maxl-algorithm}.

\setlength{\textfloatsep}{10pt}% TODO
\begin{algorithm}[th]
   \caption{MAXL algorithm}
   \label{alg:maxl-algorithm}
\begin{algorithmic}
   \STATE {\bfseries Initialize:} Network parameters: $\theta_{1}^{T}$, $\theta_{2}^{T}$; Learning rate: $\alpha$, $\beta$; Two-way image gradient loss weight: $\lambda_G$
   \WHILE{not converged}
       \FOR{each training iteration i}
            \STATE {\# sample one batch of training data}
            \STATE {$(I_{n(i)}, I_{GT(i)}) \in (I_{n}, I_{GT})$}
            \STATE {\# auxiliary-training step}
            \STATE {$\hat{I}_{c(i)}, \hat{I}_{n(i)} = f_{\theta_{1}^{T}}(I_{n(i)})$; $M = g_{\theta_{2}^{T}}(I_{n(i)})$}
            \STATE {$L = L_{Pri}(\hat{I}_{c(i)}, I_{GT(i)}) + L_{Aux}(\hat{I}_{n(i)}, I_{n(i)}, M)$}
            \STATE {Update: $\theta_{1}^{T} \leftarrow \theta_{1}^{T} - \alpha \nabla_{\theta_{1}^{T}}L$}
       \ENDFOR
       \FOR{each training iteration i}
            \STATE {\# sample one batch of training data}
            \STATE ($I_{n(i)}$, $I_{GT(i)}$) $\in$ ($I_{n}$, $I_{GT}$)
            \STATE {\# retain training computational graph}
            \STATE {$\hat{I}_{c(i)}, \hat{I}_{n(i)} = f_{\theta_{1}^{T}}(I_{n(i)})$; $M = g_{\theta_{2}^{T}}(I_{n(i)})$}
            \STATE {$L = L_{Pri}(\hat{I}_{c(i)}, I_{GT(i)}) + L_{Aux}(\hat{I}_{n(i)}, I_{n(i)}, M)$}
            \STATE {$\theta_{1}^{T+} = \theta_{1}^{T} - \alpha \nabla_{\theta_{1}^{T}}L$; $\hat{I}_{c(i)}, \_ = f_{\theta_{1}^{T+}}(I_{n(i)})$}
            \STATE {\# meta-training step}
            \STATE {Update: $\theta_2^{T} \leftarrow \theta_2^{T} - \beta \nabla_{\theta_2^{T}} (L_{Pri}(\hat{I}_{c(i)}, I_{GT(i)}) +$}
            \begin{ALC@g}
                $\lambda_G L_G(M, I_n))$
            \end{ALC@g}
       \ENDFOR
   \ENDWHILE
\end{algorithmic}
\end{algorithm}

By training the network using meta-auxiliary learning and an external dataset, the multi-task network $f_{\theta_1}$ will have representation that can generalize to various noises. 
In addition, the mask generation network also produces a mask that when used by the auxiliary loss to train the multi-task network $f_{\theta_1}$ will improve the generalization in the primary task.
We also validate the necessity of the pre-training stage in Section \ref{subsec:ablation-study}.
However, our goal is to enable the network to learn through internal training (i.e. image specific learning) by enabling test-time adaptation using meta-transfer learning (Section \ref{sec:meta-transfer-learning}) and meta-test (Section \ref{sec:meta-test}).

\subsection{Meta-transfer Learning (MTL)}
\label{sec:meta-transfer-learning}
In this step, we fine-tune the pre-train network using two different real-noise datasets to enable test-time adaptation of the network.
We use MAML-based \cite{MAML_finn} algorithm for the fine-tuning shown in Algorithm \ref{alg:meta-transfer-algorithm}.
On the inner-loop of the meta-learning, we only update the primary head and auxiliary head of the multi-task network $f_{\theta_1}$.
Meanwhile, on the outer-loop of the meta-learning stage, all the networks parameter will be updated, including the network body.
The multi-task network will be updated with the gradient from the primary objective $L_{Pri}$. 
In addition, the mask generation network is also updated with the gradient from primary loss to make the network produce a better mask.
We use unbiased sampling in our method because our goal is to make the denoising performance improve when trained with the self-supervised auxiliary loss on any examples.
This means our method can be seen as zero-shot meta-learning (no training sample) and using task-related sampling will hinder our goal to achieve generalization on any examples.

\begin{algorithm}[th]
    \caption{Meta-transfer learning}
    \label{alg:meta-transfer-algorithm}
    \begin{algorithmic}
        \STATE {\bfseries Input:} $\theta_{1}^{T}$, $\theta_{2}^{T}$; dataset $D = {D_1, D_2}$; number of inner-gradient update $K$; Auxiliary loss weight: $\lambda_{in}, \lambda_{out}$
        \STATE {\bfseries Initialize:} Learning rate: $\alpha$, $\beta$; $\theta_{1}^{T}$, $\theta_{2}^{T}$ as $\theta_{1}$, $\theta_{2}$ 
        \STATE $\theta_n = \{\theta_1^{Pri},\theta_1^{Aux}\}$
        \WHILE {not done}
            \STATE {Sample $N$ datapoints from $\mathcal{D}$, $\mathcal{B} = \{I_{n}, I_{GT}\}$}
            \FOR{each sample $j$ in $\mathcal{B}$}
                \STATE {Initialize $\theta_n^{\prime} = \theta_n$}
                \FOR{$k$ in $K$}
                \STATE {$\hat{I}_{c(j)}, \hat{I}_{n(j)} = f_{\theta_{1}^{\prime}}(I_{n(j)})$; $M = g_{\theta_2^{\prime}}({I_{n(j)}})$}
                \STATE {Compute adapted parameter of $\theta_n^{\prime}$:}
                    \begin{ALC@g}
                        \STATE {$\theta_n^{\prime} = \theta_n^{\prime} - \alpha \nabla_{\theta_n^{\prime}} \lambda_{in} L_{Aux}(\hat{I}_{n(j)}, I_{n(j)}, M)$}
                    \end{ALC@g}
                    % \STATE {Update: }
                    % \begin{ALC@g}
                    %     \STATE {$\theta_{1}^{Aux} = \theta_{1}^{Aux} - \alpha \nabla_{\theta_1} L_{Aux}(\hat{I}_{n(j)}, I_{n(j)}, M)$}
                    % \end{ALC@g}
                \ENDFOR
                \STATE {Evaluate: $\hat{I}_{c(j)}, \hat{I}_{n(j)} = f_{\{\theta_1^{Body},\theta_{n}^{\prime}\}}(I_{n(j)})$}
            \ENDFOR
            \STATE {Update $\theta_{1}$ using primary loss:}
            \begin{ALC@g}
                \STATE {$\theta_{1} \leftarrow \theta_{1} - \beta \nabla_{\theta_{1}} \lambda_{out} \sum L_{Pri}(\hat{I}_{c(j)}, I_{GT(j)})$ for each sample in $\mathcal{B}$}
            \end{ALC@g}
            
            \STATE {Update $\theta_{2}$ using primary loss:}
            \begin{ALC@g}
                \STATE {$\theta_{2} \leftarrow \theta_{2} - \beta \nabla_{\theta_{2}} \lambda_{out} \sum L_{Pri}(\hat{I}_{c(j)}, I_{GT(j)})$}
                \STATE{for each sample in $\mathcal{B}$}
            \end{ALC@g}
        \ENDWHILE
        \STATE {\bfseries Output:} $\theta_1$, $\theta_2$ as $\theta_1^{M}$, $\theta_2^{M}$
    \end{algorithmic}
\end{algorithm}

\subsection{Meta-test}
\label{sec:meta-test}

\begin{algorithm}[th]
    \caption{Meta-test}
    \label{alg:meta-test}
    \begin{algorithmic}
        \STATE {\textbf{Input:} Test dataset $\mathcal{D}_{Test}$; Network parameters: $\theta_{1}^{M}, \theta_{2}^{M}$}
        \STATE {\textbf{Initialize:} Learning rate: $\alpha$; $\theta_{1}^{M}, \theta_{2}^{M}$ as $\theta_{1}, \theta_{2}$}
        \FOR {each noisy image $I_{n}$ in $\mathcal{D}_{Test}$}
            \STATE{Initialize: $\theta_{n} = \{\theta_{1}^{Pri}, \theta_{1}^{Aux}\}$}
            \FOR {$K$ steps}
                \STATE{$M = g_{\theta_2}(I_{n})$; $\hat{I}_{c}, \hat{I}_{n} = f_{\theta_{1}}(I_{n})$}
                \STATE{Update: $\theta_{n} \leftarrow \theta_{n} - \alpha \nabla_{\theta_{n}}(L_{Aux}(\hat{I}_{n}, I_{n}, M))$}
            \ENDFOR
        \ENDFOR
        \STATE {\textbf{Output:} $\hat{I}_{c}$ for each noisy image $I_{n}$ in $\mathcal{D}_{Test}$}
    \end{algorithmic}
\end{algorithm}
\vskip 0.1in

Algorithm \ref{alg:meta-test} shows how meta-test is being done in the testing stage for fast test-time adaptation. 
The network body of the multi-task network is frozen at this stage. 
In the meta-testing stage, given a noisy image $I_{n}$, we adapt the primary and auxiliary head to denoise this image by using $K$ gradient steps.
We use the auxiliary loss $L_{Aux}$ as the objective to adapt the network parameters in the testing stage since we have encouraged the multi-task network to improve the primary task performance when trained with this loss.

\section{Experiments}

\subsection{Implementation Details}
In the pre-training stage, we train our network with the MAXL algorithm by using DIV2K \cite{agustsson2017ntire} dataset with synthetic degradation consisting of salt-and-pepper, gaussian, and speckle noise.
We set $\alpha$ and $\beta$ in this stage as $0.001$ and optimize the network using Adam optimizer.
For the two-way image gradient loss weight ($\lambda_G$), we use $\lambda_G = 0.01$.
After pre-training, we further fine-tune our network using the meta-transfer learning algorithm.
For this stage, we use SIDD \cite{abdelhamed2018high} following \cite{zamir2020learning} especially the small version consisting of 160 noisy-clean pairs and Poly \cite{xu2018real} dataset. 
Similar to the pre-training, we use Adam as the optimizer and set $\alpha=\beta=0.00005$ and $K=5$.
Furthermore, we set auxiliary loss weight for the inner and outer update as $\lambda_{in}=\lambda_{out}=10$.
For the meta-testing, we use smaller $\alpha$ than the meta-transfer learning $\alpha=0.00001$.
In pre-training and fine-tuning, we use a batch size of $16$ and $2$ respectively with a patch size of $128\times128$.
For the evaluation, we conduct the validation using the validation split of the training dataset while using Nam \cite{nam2016holistic} as the testing dataset.
We use PSNR and SSIM as the evaluation metric.
In terms of PSNR, the improvement of 0.05 and 1 dB can be considered as a contribution and significant contribution respectively.
Meanwhile, an improvement of 0.01 in terms of SSIM can be considered as a contribution.

\subsection{Pre-training Results}
\label{subsec:pretraining-results}

\begin{table*}[t]
\caption{The results of pre-training using different multi-task network architecture (top row) with reconstruction loss as the auxiliary loss $L_{Aux} = L_{Rec}$. We also investigate the results of using different auxiliary losses with MAXL (bottom row) using our multi-task network architecture. Aux stands for auxiliary. All of the experiments are trained with the primary loss $L_{Pri}$ and the specified auxiliary loss $L_{Aux}$.}
\label{table:pretraining-results}
\vskip 0.15in
\begin{center}
\begin{small}
\begin{sc}
\resizebox{0.68\textwidth}{!}{
    \begin{tabular}{lc|cc|c}
    \toprule
    \multirow{2}{*}{Details} & \multicolumn{2}{c}{Validation} & \multicolumn{2}{c}{Testing} \\
     & PSNR & SSIM & PSNR & SSIM \\
    \midrule
    Our Architecture + $L_{Aux} = L_{Rec}$ & \textbf{31.1540} & \textbf{0.8711} & \textbf{35.7237} & 0.9054 \\
    \cite{chi2021test} + $L_{Aux} = L_{Rec}$ & 30.8964 & 0.8664 & 33.4407 & \textbf{0.9153} \\
    \midrule
    MAXL + $L_{Aux} = L_{Rec}$ & 31.1417 & 0.8691 & 33.4551 & 0.9207\\
    MAXL + $L_{Aux} = L_{MaskRec}$ & \textbf{31.3130} & \textbf{0.8719} & 35.8208 & 0.9044 \\
    MAXL + $L_{Aux} = L_{MaskRec}$ + $L_{G}$ (Ours) & 31.2193 & 0.8634 & \textbf{36.1182} & \textbf{0.9050} \\
    \bottomrule
    \end{tabular}}
\end{sc}
\end{small}
\end{center}
\vskip -0.18in
\end{table*}

% TODO: Settings
In this experiment, we conduct two different experiments to demonstrate our contribution. 
We train the network using DIV2K dataset with synthetic degradation. 
For the evaluation, we conduct the validation using validation images from DIV2K synthetic degradation, while testing using images from the Nam dataset with real noise.
Quantitative results in terms of PSNR and SSIM metrics can be seen in Table \ref{table:pretraining-results}.

In the first experiment, we compare two different architectures of the multi-task network $f_{\theta_1}$ to evaluate which network will get more benefits when trained using auxiliary loss.
The first architecture is our proposed architecture which uses a sequential design where the auxiliary head is placed after the primary head.
Meanwhile, the second architecture is the architecture from \cite{chi2021test} which is a single feature extractor with two parallel heads (primary \& auxiliary). 
The details of this architecture can be seen in Figure \ref{fig:multi-task-baseline} in the supplementary material.
Both networks have a similar number of parameters.
To train the network, we use the primary loss and change the auxiliary loss to reconstruction loss $L_{Rec}$.
We change the loss to measure the capability of the multi-task network in an isolated manner without any effect from the mask generation network $g_{\theta_2}$.
The results can be seen in the top row of Table \ref{table:pretraining-results} where our proposed architecture outperforms the other architectures by 0.3 dB on validation and 2.3 dB on the testing PSNR.
Based on the SSIM metric, our architecture also achieves better validation SSIM but lower testing SSIM.
This shows that our architecture achieves better generalization when trained with auxiliary loss compared to the baseline especially in the case of unseen real noise.
The worse results of the network architecture from \cite{chi2021test} may be due to the placement of the auxiliary head that can hurt the performance when placed after the feature extractor i.e. negative transfer issue problem.

After validating that our proposed architecture is a better option when trained with an auxiliary loss $L_{Aux}$, we compare different auxiliary loss functions trained with MAXL.
% When trained without MAXL (middle row of Table \ref{table:pretraining-results}), we can see that using our proposed masked reconstruction loss $L_{MaskRec}$ achieves better validation and testing PSNR.
% This shows the benefit of using the proposed loss even though the validation and testing SSIM score is slightly lower than other losses.
% We observe the collapse situation where the mask becomes zero masks which may contribute to the worse SSIM score when the network is trained with masked reconstruction loss.
% Furthermore, by using auxiliary loss, the generalization performance can greatly improve where using auxiliary loss in the form of reconstruction loss $L_{Rec}$ can improve the testing performance by 1.8 dB, and using masked reconstruction loss further improves the testing PSNR by 0.3 dB.
% This result denotes the importance of using auxiliary loss to achieve better generalization on unseen noise.
When trained with MAXL (bottom row of Table \ref{table:pretraining-results}), the validation and testing performance of using masked reconstruction loss ($L_{MaskRec}$) are consistently better compared to using reconstruction loss ($L_{Rec}$).
Since using MAXL also cannot prevent the collapse situation, we use two-way image gradient loss ($L_{G}$) to regularize the mask thus preventing the collapse situation.
This loss further improves the generalization to unseen noise by improving both PSNR and SSIM scores but decreases the validation performance.
Since our goal is to improve the generalization of multi-task networks on unseen real noise, we use the network that achieves the best testing performance to be fine-tuned using the meta-transfer learning algorithm.

\subsection{Meta-transfer Learning Results}

\begin{table*}[t]
\caption{The results of meta-transfer learning using the proposed algorithm compared with other SOTA methods.}
\label{table:meta-transfer-learning-result}
\vskip 0.15in
\begin{center}
\begin{small}
\begin{sc}
\resizebox{0.68\textwidth}{!}{
    \begin{tabular}{lc|cc|cc}
    \toprule
    \multirow{2}{*}{Details} & \multicolumn{2}{c}{Validation} & \multicolumn{2}{c}{Testing} & Number of \\
     & PSNR & SSIM & PSNR & SSIM & Parameters \\
    \midrule
    Ours without Meta-testing & 41.4792 & 0.9633 & 39.2499 & 0.9672 & \textbf{0.66 M} \\
    Ours with Meta-testing & \textbf{41.5086} & \textbf{0.9636} & \textbf{39.2653} & 0.9685 & \textbf{0.66 M} \\
    RIDNet \cite{anwar2019real} & 40.2185 & 0.9497 & 38.2214 & 0.9621 & 1.49 M \\
    MIRNet \cite{zamir2020learning} & 41.0460 & 0.9609 & 38.9807 & \textbf{0.9705} & 31.79 M \\
    \bottomrule
    \end{tabular}}
\end{sc}
\end{small}
\end{center}
\vskip -0.18in
\end{table*}

\begin{figure*}[ht]
\vskip 0.1in
\begin{center}
    \centerline{\includegraphics[width=0.68\textwidth]{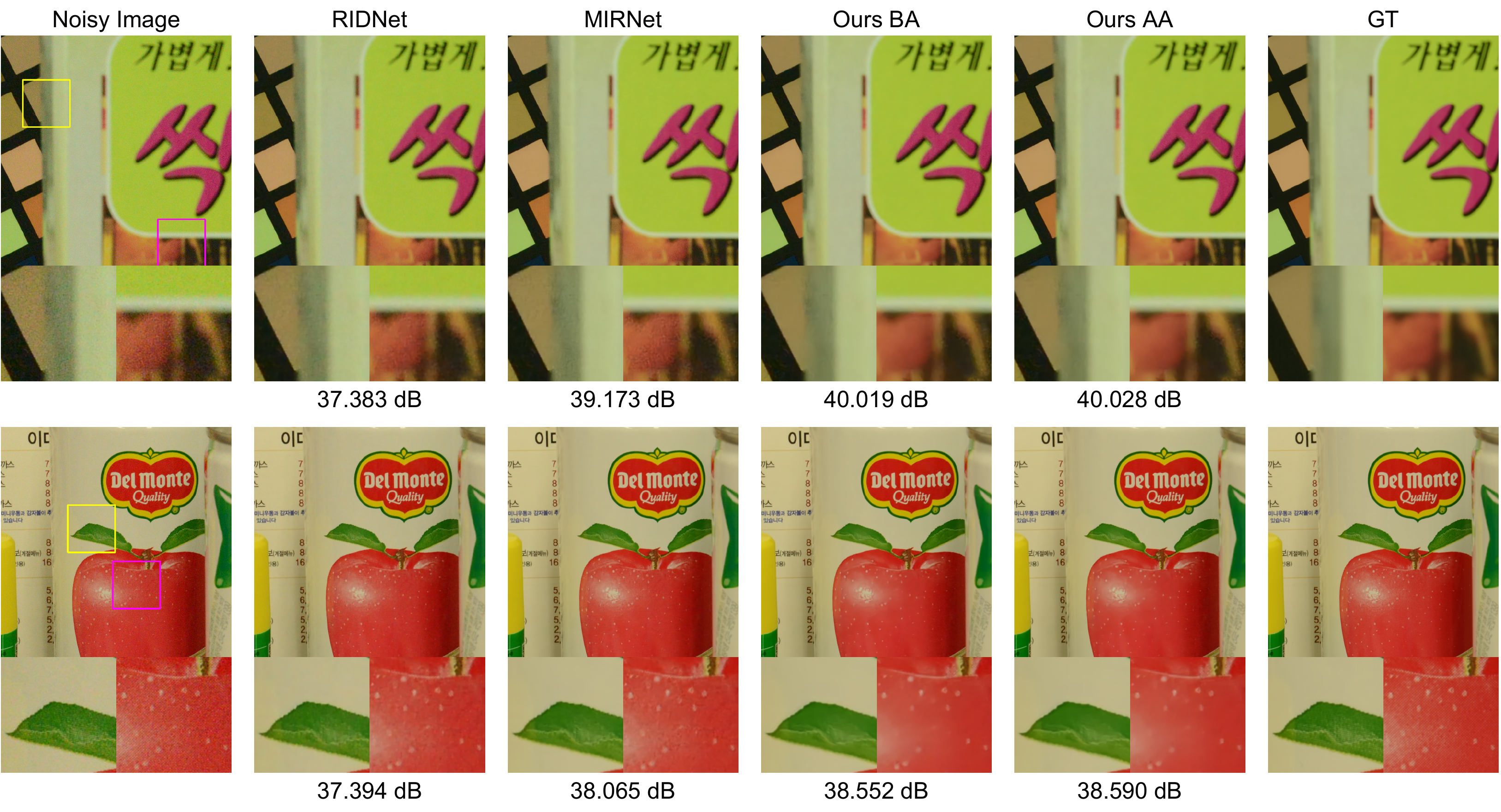}}
    \caption{Qualitative results of our method compared to others. BA and AA stand for before adaptation and after adaptation respectively,  where the meta-testing is applied on each example of the After Adaptation (AA) result.}
    \label{fig:qualitative-results}
    \end{center}
    \vskip -0.3in
\end{figure*}

In this experiment, we fine-tune the best pre-trained network in Section \ref{subsec:pretraining-results} using the meta-transfer learning algorithm and show the performance before adaptation (without meta-testing) and after adaptation (with meta-testing).
We also compare the result with the recent SOTA methods: RIDNet \cite{anwar2019real} and MIRNet \cite{zamir2020learning}.
We cannot compare our method with a similar method in denoising task \cite{lee_2020} since there is no official code from this method and this method is not designed to handle real noise.
In addition, we only choose the best SOTA which can be run in our machine where MIRNet is the SOTA that achieves the fourth rank in denoising task.
We use the dataset described in the implementation details of the meta-transfer learning stage to train all of the methods.
Due to the limitation of our machine, we can only conduct the meta-transfer learning using a patch size of $1024$ which makes the evaluation (validation and testing) is also conducted using a center crop with the size of $1024$.
In addition, we also try various hyperparameters but cannot find any meaningful improvement besides the one that we use.

Quantitative comparison can be seen in Table \ref{table:meta-transfer-learning-result}.
The results show that using meta-testing to adapt the network for each image in the evaluation dataset can consistently improve the validation performance and testing performance.
In addition, we can see that our method achieve the best result compared to SOTA methods.
Compared to the MIRNet result, our method with meta-testing improves the validation performance by 0.46 dB and 0.0027 in terms of PSNR and SSIM respectively.
The testing performance in terms of PSNR also improves by 0.28 dB.
Interestingly, even though the number of parameters of our network is very small compared to other SOTAs ($\sim$2x and 48x compared to RIDNet and MIRNet respectively), we can still outperform the SOTA method even though the SSIM score is slightly lower compared to MIRNet.
This shows a promising research direction where we can pursue an improvement in denoising tasks by modifying the training algorithm where the current trend of the SOTA methods is dominated by the improvement in the design of network architecture.
The reason for slight performance gain after adaptation (0.03 dB) is likely due to the small learning capacity of the network where the number of parameters of the adapted head is only 0.12 M which can be considered small.
We conjecture that better improvement can be observed if we increase the number of parameters by making the larger network.
However, due to the limitation of our machine, we cannot experiment using a larger network because of the multi-gradient computation.
We conduct an additional feature visualization study (Section \ref{app:sec-feature-map-visualization} in the supplementary material) which shows a high difference in feature maps after adaptation.

In terms of qualitative comparison, the visual comparison on challenging examples can be seen in Figure \ref{fig:qualitative-results}.
The first example (top row) shows that both of the baselines fail to effectively remove the real noise shown in the left patch and right patch of the example.
In addition, some leftover artifacts can be seen especially in the homogeneous region.
Our method successfully denoises the real noise and maintains the smoothness of the homogeneous region without any artifacts. 
In the second example (bottom row), all of the methods fail to recover fine textural details of paper texture, especially in the right patch.
However, our method successfully denoises the real noise and produces visually pleasing images in the left patch and right patch of the example.
The other baseline methods fail to denoise real noise but interestingly try to maintain fine textural details (e.g. paper texture) which makes both methods produce noisy artifacts that do not look visually pleasing.

\subsection{Ablation Study}
\label{subsec:ablation-study}

\begin{table*}[t]
\caption{The results of the ablation study. Top row: the results of fine-tuning with meta-transfer learning (MTL) using different losses on our MAXL pre-trained network. Middle row: the results of fine-tuning using MTL on a Randomly Initialized (RI) network. Bottom row: the results of updating all of the network parameters instead of primary and auxiliary heads only. All of the results are the result after running meta-testing.}
\label{table:ablation-study}
\vskip 0.15in
\begin{center}
\begin{small}
\begin{sc}
\resizebox{0.68\textwidth}{!}{
    \begin{tabular}{lc|cc|c}
    \toprule
    \multirow{2}{*}{Details} & \multicolumn{2}{c}{Validation} & \multicolumn{2}{c}{Testing} \\
     & PSNR & SSIM & PSNR & SSIM \\
    \midrule
    MAXL + MTL with Aux Loss (MRL) & \textbf{41.5086} & \textbf{0.9636} & \textbf{39.2653} & \textbf{0.9685} \\
    MAXL + MTL with Aux Loss (RL) & 41.4466 & 0.9634 & 39.1280 & 0.9683 \\
    \midrule
    RI + MTL with Aux Loss (MRL) & 40.1913 & 0.9554 & \textbf{38.1708} & \textbf{0.9628} \\
    RI + MTL with Aux Loss (RL) & \textbf{40.2040} & \textbf{0.9570} & 38.0766 & 0.9621 \\
    \midrule
    Updating Body + Head: MAXL + MTL with Aux Loss (MRL) & \textbf{41.4919} & \textbf{0.9632} & \textbf{39.1973} & \textbf{0.9691} \\
    Updating Body + Head: RI + MTL with Aux Loss (MRL) & 40.0701 & 0.9536 & 37.6599 & 0.9553 \\
    \bottomrule
    \end{tabular}}
\end{sc}
\end{small}
\end{center}
\vskip -0.18in
\end{table*}

In this study, we investigate the impact of our proposed method such as the impact of MAXL pre-training, masked reconstruction loss, and updating only the head of the multi-task network.
The top row and middle row of Table \ref{table:ablation-study} shows that training with masked reconstruction loss objective consistently outperforms reconstruction loss, especially on the testing performance.
The proposed masked reconstruction loss can improve the testing performance around 0.1 dB which shows that the network can achieve better generalization when trained using this objective, especially when coupled with the meta-transfer learning.
Different from the results of training using MAXL, we observe no collapse situation throughout fine-tuning both on MAXL pre-trained network and randomly initialized network.
Figure \ref{fig:mask-maxl-mtl} and Figure \ref{fig:mask-scratch-mtl} in the supplementary material show the evolution of the mask through training where the mask constantly evolves in early training then becomes constant when the multi-task network starts to converge (i.e. after epoch 100).
This shows the benefit of masked reconstruction loss that can provide the benefit of choosing the certain region that needs to be reconstructed so it can improve the performance of the primary task in different stages of learning.

To study the impact of pre-training, the result in the top row and middle row of Table \ref{table:ablation-study} can be compared.
The results show that pre-training is indeed required to achieve better performance both on validation and testing.
Note that the performance gap is not caused by the convergence issue since both settings (fine-tune on the pre-trained network and randomly initialized network) are already converged.
The bottom row of Table \ref{table:ablation-study} again consolidates this fact where the gap between the result of fine-tuning on the pre-trained network compared to the randomly initialized network is around 1.5 dB both on validation and testing performance.

The last study is to investigate the impact of updating only the multi-task network's head in the inner loop of the fine-tuning stage.
The result can be seen in the top row and bottom row of Table \ref{table:ablation-study}.
Validation performance of updating only head compared to updating all of the network parameters is similar.
However, the testing performance in terms of PSNR can be improved by 0.07 even though the SSIM scores slightly drop.
This denotes that both updating only head and all of the network parameters in the inner loop of fine-tuning achieve similar results.
Similar observation also can be seen in \cite{raghu2020rapid}.

\section{Conclusion}
In this paper, we propose a combination of algorithms to enable test-time adaptation on the problem of real image denoising. 
We first design a network consisting of a multi-task and mask generation network. 
Then, we propose a novel self-supervised masked reconstruction loss as the auxiliary loss to train the network.
To train the network, we propose to use two-stage learning.
The first stage pre-train the network using a meta-auxiliary learning algorithm to get better meta-initialization. 
Meanwhile, the second stage further fine-tunes the network using meta-transfer learning.
This combination of meta-auxiliary learning and meta-transfer learning improves the generalization performance of the network against various unseen noise and enables test-time adaptation.
The adaptation makes the network can adapt to real noisy images within a few gradient updates.
Various experiments show the contribution of the components of our method and also show that our method can outperform other SOTA methods.
Yet, we still find it is necessary to conduct more extensive experiments on other real-noise datasets to validate the proposed method and use a larger version of our network to further validate our contribution.


\bibliography{example_paper}
\bibliographystyle{icml2021}
%Additional command
\clearpage


%%%%%%%%%%%%%%%%%%%%%%%%%%%%%%%%%%%%%%%%%%%%%%%%%%%%%%%%%%%%%%%%%%%%%%%%%%%%%%%
%%%%%%%%%%%%%%%%%%%%%%%%%%%%%%%%%%%%%%%%%%%%%%%%%%%%%%%%%%%%%%%%%%%%%%%%%%%%%%%
\newpage
\appendix
%TODO: Complete supplementary material

\section*{Supplementary Material}
In the following, we provide additional details about the experiments (Section \ref{sec:additional-experiment-details}).
The experiment details consist of the baseline architecture of the multi-task network and additional implementation details.
In addition, we also provide additional experiments to give some visualizations about each component of our method (Section \ref{sec:additional-experiments}). 
We conduct three additional experiments consisting of mask visualization through different epochs of meta-transfer learning, unfolding adaptation process in some examples, and feature map visualization to compare the feature map before and after adaptation.

\section{Experiment Details}
\label{sec:additional-experiment-details}

\subsection{The Details of Multi-task Network Architecture Baseline}

\begin{figure}[ht]
\vskip 0.1in
\begin{center}
    \centerline{\includegraphics[width=\columnwidth]{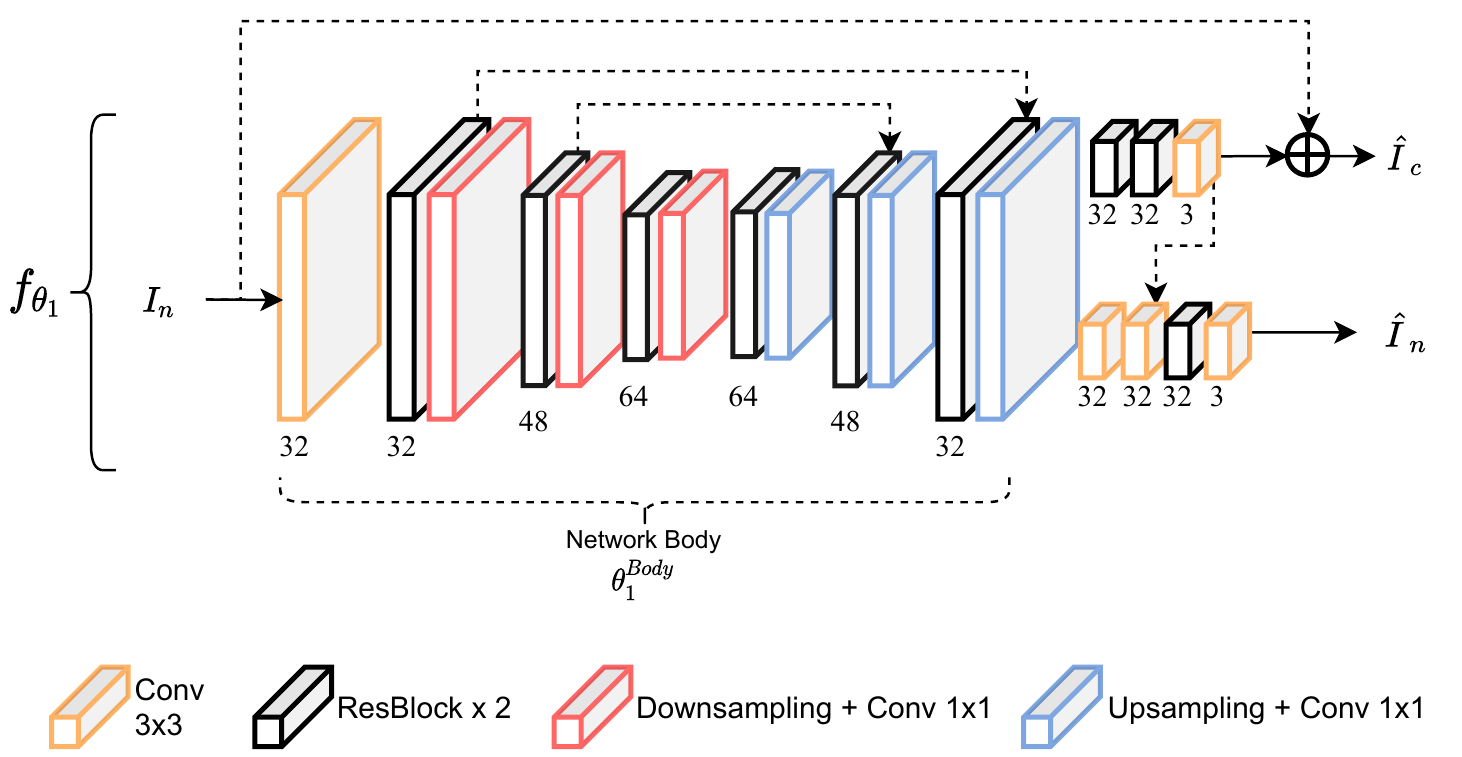}}
    \caption{Multi-task network architecture from \cite{chi2021test}.}
    \label{fig:multi-task-baseline}
    \end{center}
    \vskip -0.2in
\end{figure}

The architecture in Figure \ref{fig:multi-task-baseline} uses a single feature extractor with multi parallel head.
Note that this architecture is the same as the network architecture in \cite{chi2021test} but we modify the number of features for each convolutional layer. 
In addition, we also do not use their multi-scale design with feature recurrence for better comparison with our multi-task network architecture.

\subsection{Additional Implementation Details}
Some omitted experiment details can be seen in the following points:
\begin{itemize}
    \item Synthetic noise degradation: we use three different synthetic noises consisting of: 1) gaussian noise with a standard deviation between 5 until 50 in pixel-scale (0-255), 2) speckle noise with a standard deviation between 5 until 50 in pixel-scale, and 3) salt-and-pepper with total amount of $\mathcal{U}(0, 0.01)$ and salt probability of $\mathcal{U}(0.3, 0.8)$. We apply each degradation in a random sequence.
    \item GPU: we use a single NVIDIA Titan Xp in all experiments.
    \item MIRNet and RIDNet implementation: we use the official code provided by the author where we take their network code and train it using our pipeline. 
\end{itemize}

\section{Additional Experiments}
\label{sec:additional-experiments}

\subsection{Mask Visualization in Meta-Transfer Learning Stage}

\begin{figure}[ht]
\vskip 0.1in
\begin{center}
    \centerline{\includegraphics[width=\columnwidth]{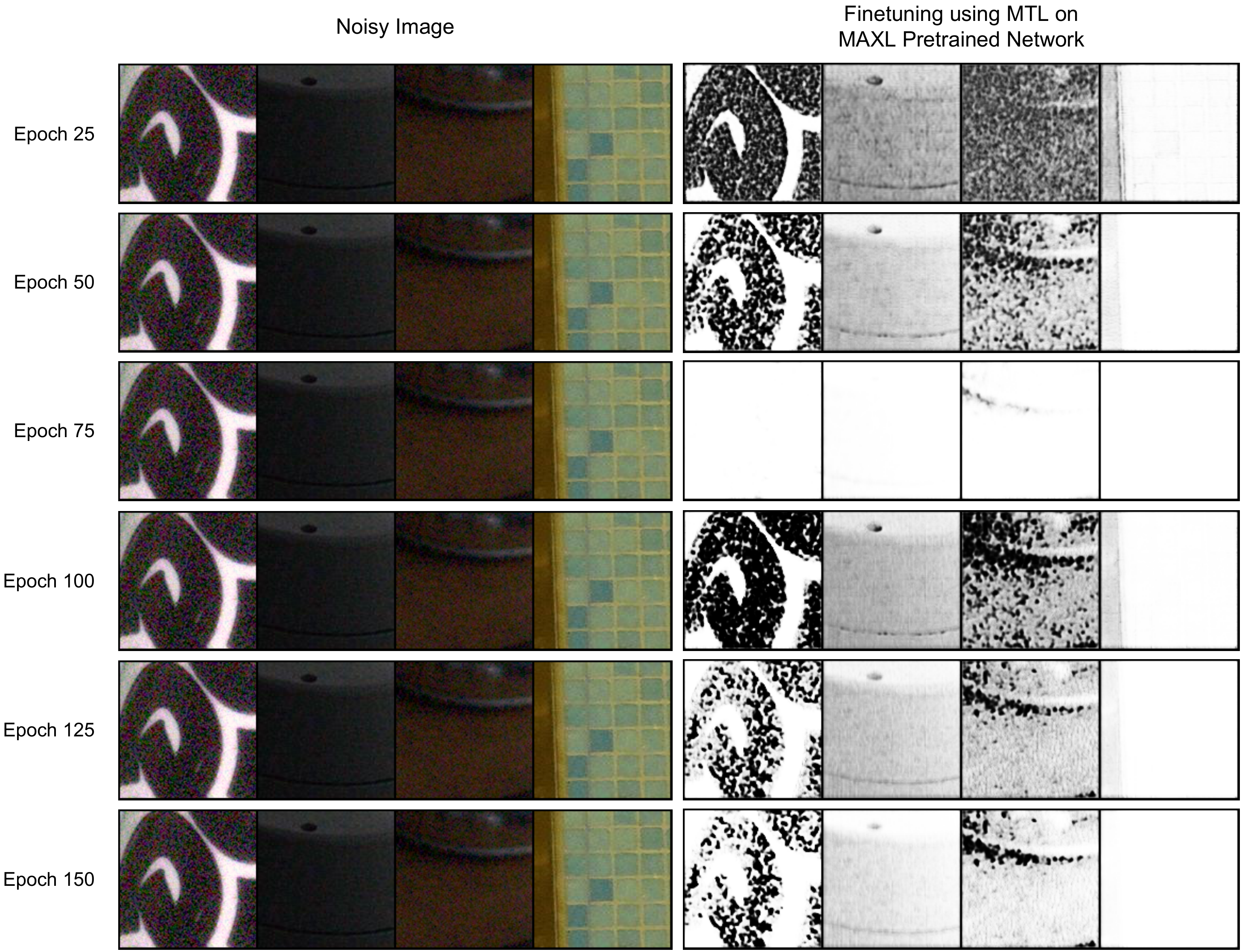}}
    \caption{Visualization of the mask through different fine-tuning epochs using meta-learning on MAXL pre-trained network.}
    \label{fig:mask-maxl-mtl}
    \end{center}
    \vskip -0.2in
\end{figure}

\begin{figure}[ht]
\vskip 0.1in
\begin{center}
    \centerline{\includegraphics[width=\columnwidth]{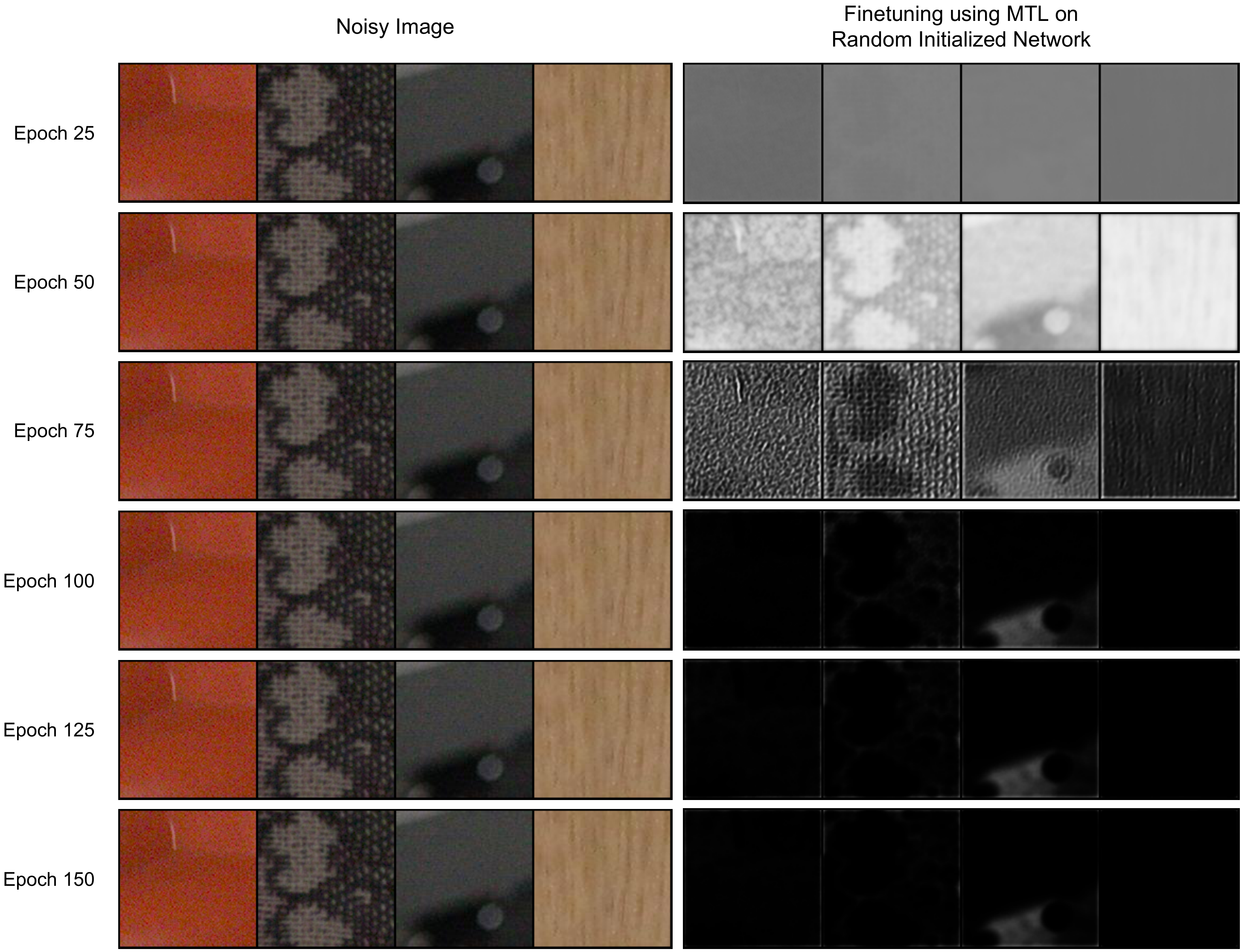}}
    \caption{Visualization of the mask through different fine-tuning epochs using meta-learning on the randomly initialized network.}
    \label{fig:mask-scratch-mtl}
    \end{center}
    \vskip -0.2in
\end{figure}

\begin{figure*}[ht]
\vskip 0.1in
\begin{center}
    \centerline{\includegraphics[width=0.8\textwidth]{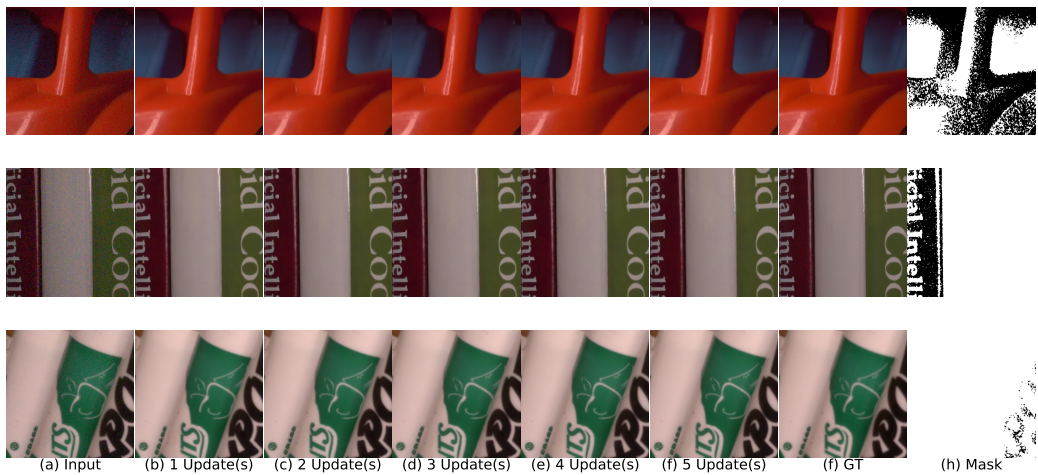}}
    \caption{Visual illustration of the unfolded adaptation process in the meta-testing with K=5.}
    \label{fig:unfolding-results}
    \end{center}
    \vskip -0.2in
\end{figure*}

Figure \ref{fig:mask-maxl-mtl} and Figure \ref{fig:mask-scratch-mtl} show the evolution of the mask produced by mask generation network through different fine-tuning epochs. 
The mask always changes through different fine-tuning epochs where the MAXL pre-trained mask generation network does not produce any meaningful changes on the mask after epoch 125.
Meanwhile, the randomly initialized mask generation network stops producing any meaningful changes on the generated mask after epoch 100.
These results show that the mask generation network achieves faster convergence when initialized randomly which may denote that MAXL pre-trained network has better robustness to the local optimum.

Both Figure \ref{fig:mask-maxl-mtl} and Figure \ref{fig:mask-scratch-mtl} also show that the mask focus more on the noisy part and large homogeneous region of the image through the different stage of learning before converging.
This aligns with our goal where we want to make the auxiliary loss focus more on the noisy part of the image since it can help the multi-task network to differentiate which part of the image is the real noise or which part of the image is the important details.
In addition, the evolution of the mask shows the benefit of the mask reconstruction loss which can help the training of multi-task network in each epoch by providing different region that needs to be reconstructed so the primary task performance can be improved. 
Yet, we also notice that our method have some weakness when the mask generation network fails to generalize across different real noisy images.
This condition can reduce the performance of the method after adaptation.

\subsection{Unfolding Adaptation Process}

Figure \ref{fig:unfolding-results} shows the result of the predicted clean image within every inner loop iteration of the meta-testing. 
Results show how the input noisy image is immediately cleaned with a single iteration and as the adaptation progress, more noisy area become cleaner.
In addition, even though it is only marginal, we can observe some region that becomes sharper as the adaptation progress.

\subsection{Feature Map Visualization}
\label{app:sec-feature-map-visualization}

\begin{figure*}[ht]
\vskip 0.1in
\begin{center}
    \centerline{\includegraphics[width=0.7\textwidth]{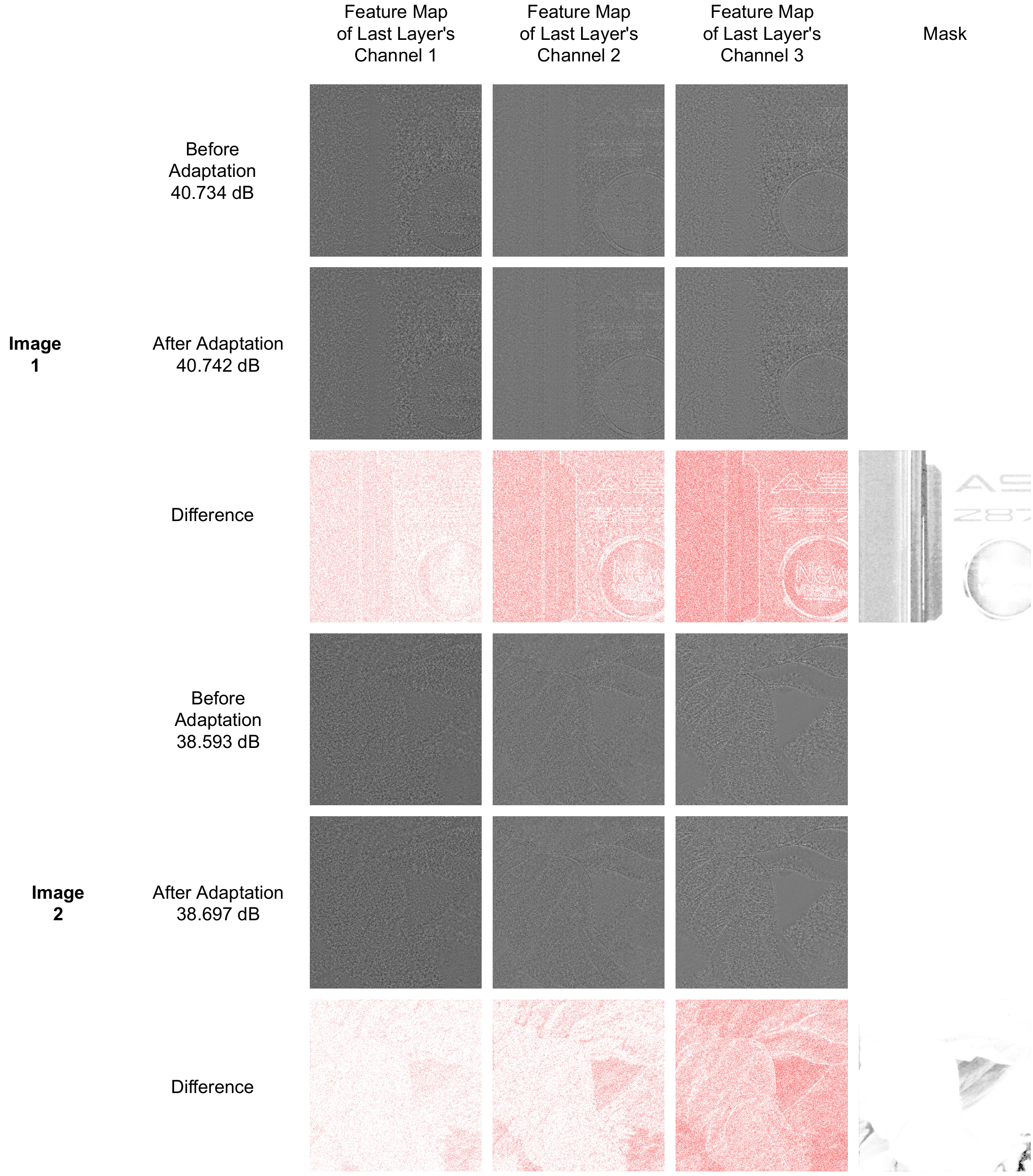}}
    \caption{Feature map visualization of primary head's last layer.}
    \label{fig:feature-map-visualization}
    \end{center}
    \vskip -0.2in
\end{figure*}

Figure \ref{fig:feature-map-visualization} shows the visualization of the last layer's feature map of the multi-task network's primary head.
This visualization shows the difference in feature maps after the adaptation of the multi-task network.
As can be seen, the difference of the feature map in each channel before and after adaptation is large for both the marginal improvement in image 1 and the large improvement in image 2.
This shows that sometimes the adaptation fails to improve the performance of the denoising task significantly.
One possible reason is that the adaptation still fails to generalize in any real noise examples.
We also show the mask of the given example and cannot get any correlation between the difference of the feature map before and after adaptation.
Interestingly, even though some region of the mask is zero (black pixel), those regions still have differences in features after adaptation.
This shows that the masked reconstruction loss can help the primary head adapt to the whole region of the input image even though the auxiliary loss is only computed in some regions (white pixels in the mask).

\end{document}